\documentclass{article} %
\usepackage{colm2024_conference}

\usepackage{amsmath,amsfonts,bm}

\def\eqref#1{equation~\ref{#1}}

\def\1{\bm{1}}

\DeclareMathAlphabet{\mathsfit}{\encodingdefault}{\sfdefault}{m}{sl}
\SetMathAlphabet{\mathsfit}{bold}{\encodingdefault}{\sfdefault}{bx}{n}

\usepackage{times}
\usepackage{latexsym}

\usepackage[utf8]{inputenc} %
\usepackage[T1]{fontenc}    %
\usepackage[hidelinks]{hyperref}       %
\usepackage{url}            %
\definecolor{darkblue}{rgb}{0, 0, 0.5}
\hypersetup{colorlinks=true, citecolor=darkblue, linkcolor=darkblue, urlcolor=darkblue}
\usepackage{booktabs}       %
\usepackage{amsfonts}       %
\usepackage{nicefrac}       %
\usepackage{microtype}      %
\usepackage{inconsolata}
\usepackage{xcolor}         %
\usepackage[pdftex]{graphicx}
\usepackage{amsmath}
\usepackage[capitalize]{cleveref}
\usepackage{xspace}
\usepackage{adjustbox}
\usepackage{wrapfig}
\usepackage{subcaption}
\usepackage{multirow}
\usepackage{caption}
\usepackage{multicol}
\usepackage{cancel}
\usepackage[compact]{titlesec}
\usepackage{changepage}
\usepackage{enumitem}
\usepackage[multiple]{footmisc}
\usepackage[textsize=tiny]{todonotes}

\newcommand{\subj}{s}

\newcommand{\subjrep}{\mathbf{\subj}}

\newcommand{\lre}{\textsc{LRE}}

\definecolor{forestgreen}{rgb}{0.13, 0.55, 0.13}

\title{Locating and Editing Factual Associations in Mamba}

\author{%
Arnab Sen Sharma\thanks{Correspondence to \href{mailto:sensharma.a@northeastern.edu}{sensharma.a@northeastern.edu}, Code available at \href{https://romba.baulab.info}{\texttt {romba.baulab.info}}},\, David Atkinson,\, and David Bau \\
Khoury College of Computer Sciences, Northeastern University
}

\colmfinalcopy %
\begin{document}
\maketitle

\begin{abstract}

We investigate the mechanisms of factual recall in the Mamba state space model.  Our work is inspired by previous findings in autoregressive transformer language models suggesting that their knowledge recall is localized to particular modules at specific token locations; we therefore ask whether factual recall in Mamba can be similarly localized. To investigate this, we conduct four lines of experiments on Mamba. First, we apply causal tracing or interchange interventions to localize key components inside Mamba that are responsible for recalling facts, revealing that specific components within middle layers show strong causal effects at the last token of the subject, while the causal effect of intervening on later layers is most pronounced at the last token of the prompt, matching previous findings on autoregressive transformers. Second, we show that rank-one model editing methods can successfully insert facts at specific locations, again resembling findings on transformer LMs. Third, we examine the linearity of Mamba's representations of factual relations.  Finally we adapt attention-knockout techniques to Mamba in order to dissect information flow during factual recall. We compare Mamba directly to a similar-sized autoregressive transformer LM and conclude that despite significant differences in architectural approach, when it comes to factual recall, the two architectures share many similarities.

\end{abstract}
\section{Introduction}
\label{sec:intro}   

Studies of autoregressive transformer language models' (LMs) processing of factual statements such as \textit{The Eiffel Tower is located in Paris}, have identified a localized pattern of internal computations when recalling facts~\citep{meng2022locating, meng2022mass, geva2023dissecting, hernandez2023linearity, Nanda2023fact}, and have further found that those LMs can be edited by making single-layer rank-one changes in model parameters to alter a specific fact. Although these localized phenomena appear to generalize across autoregressive transformer LMs, the extent to which similar locality might appear in very different architectures---such as recurrent networks (RNNs)---has not yet been investigated.

In this paper we investigate the internal mechanisms of Mamba~\citep{gu2023mamba}, a recently-proposed state-space language model, a type of RNN that achieves per-parameter performance that is competitive with transformers. Specifically, we ask whether factual recall within Mamba exhibits locality similar to the patterns observed in autoregressive transformer language models.

Our paper is a case study confronting a key methodological challenge that broadly faces interpretability researchers: as state-of-the-art neural network architectures evolve, we must ask, can the detailed analytical methods and tools developed for one neural architecture, such as transformer LMs, be generalized and applied to a different neural architecture, such as Mamba? In this paper we are able to answer the question with a qualified ``yes'': we find that many of the methods used to analyze transformers can also provide insights on Mamba. We also discuss mismatches---that is, interpretation methods (such as path-dependent attention patching) that do not transfer to Mamba as easily due to architectural constraints.

We begin by studying whether activation patching ~\citep{wang2022interpretability} can be successfully applied to Mamba. Known variously as causal mediation analysis~\citep{vig-etal-2020}, causal tracing~\citep{meng2022locating}, and interchange interventions~\citep{geiger2021inducing}, activation patching techniques can successfully identify specific model components in transformer LMs that play crucial roles in performing a task.  
We ask whether Mamba can be productively studied the same way, even though the architectural components of Mamba are very different: for example, instead of attention heads and MLP modules, Mamba is composed of convolutions, gates, and state-space modules. To answer, we adapt activation patching to Mamba, and ask if any sparsity patterns emerge which provide insights into the respective roles of its components.

We also study whether rank-one model editing can be applied to Mamba. While studies of transformers~\citep{meng2022locating,meng2022mass,hase2024does} have found that there are a range of MLP modules within which factual knowledge can be inserted by making a single rank-one change in parameters, Mamba does not have MLP modules, so we ask if there are any other modules that can be similarly edited to insert knowledge.  As with previous studies of transformers, the key question is whether factual associations can be edited with both specificity (without interfering with unrelated facts) and generalization (while remaining robust to rewordings of the edited fact).

Finally, we apply methods for understanding the overall information flows in Mamba. Inspired by the findings of~\citet{hernandez2023linearity}, we measure the linearity of the relations between subject and object embeddings. And inspired by~\citet{geva2023dissecting}, we examine information flow by adapting attention-blocking methods to the attention-free Mamba architecture.

In this work 
we conduct our experiments on Mamba-2.8b, the largest available LM in Mamba family, and for comparison we conduct the same experiments on the similarly sized Pythia-2.8b \citep{biderman2023pythia} autoregressive transformer LM.
\section{Background on Mamba}
\label{sec:background}

\label{sec:mambablock}
Mamba, introduced in \citet{gu2023mamba}, is a recent family of language models based on state space models (SSMs). SSMs are designed to model the evolution of a hidden state across time with a first-order differential equation~\citep{koopman1999statistical,durbin2012time}, and when they are used as the recurrent state of an RNN, they can enable highly efficient parallelized training~\citep{gu2021efficiently}. To achieve good performance in language modeling, the Mamba SSM introduces input-dependent parameterization or \textit{selective}-SSM instead of the traditional time-invariant SSMs. Mamba uses a special architecture called \textbf{MambaBlock}\footnote{In their paper, \cite{gu2023mamba} call this component Mamba---the same name as the LM family.}, which is stacked homogeneously, replacing both attention and MLP blocks used in transformer layers. Here, we focus on the different operations performed inside a MambaBlock. 

\begin{wrapfigure}{R}{0.55\textwidth}
\vspace{-22pt}
    \centering
    \includegraphics[width=0.55\textwidth]{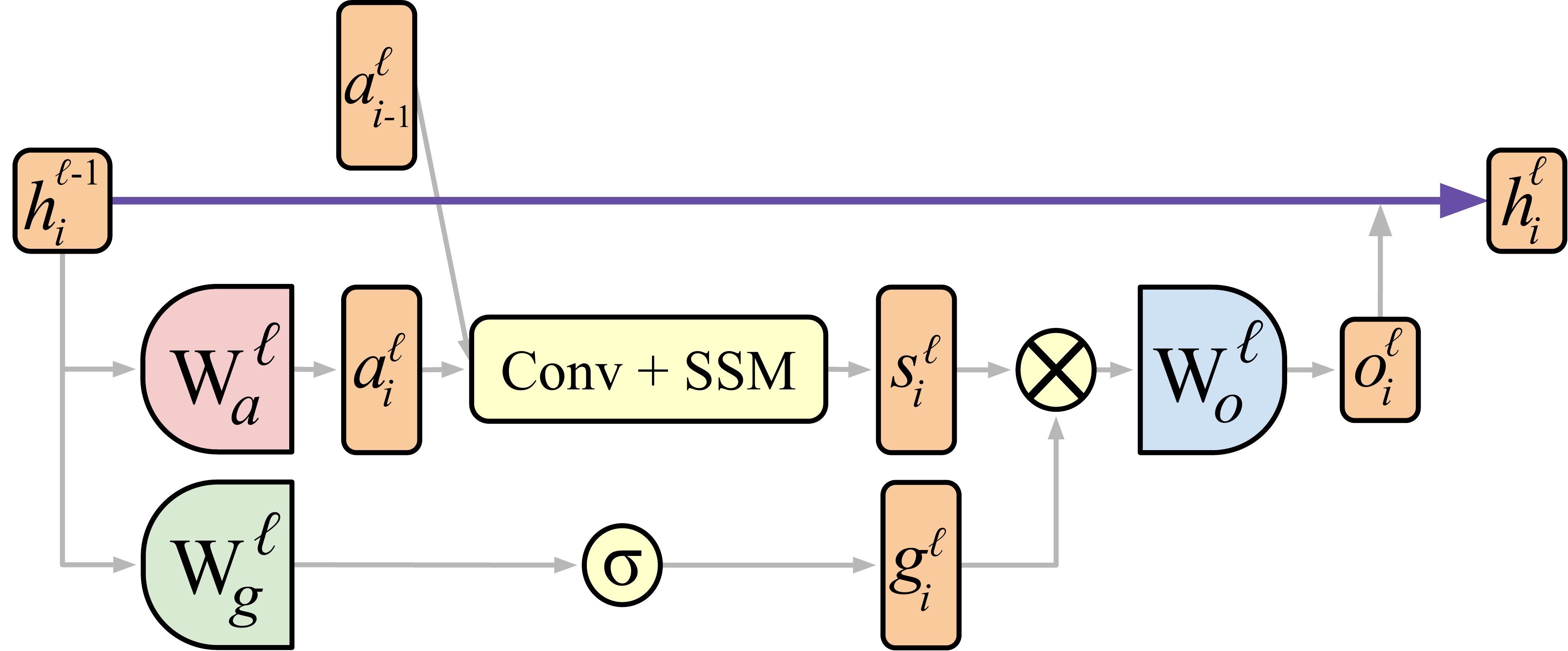}
    \caption{\small Architecture of a MambaBlock. Projection matrices $\text{W}_a^{\ell}$ and $\text{W}_g^{\ell}$ have the shape $2d \times d$, while $\text{W}_o^{\ell}$ has the shape $d \times 2d$. $h, a, g, s, \;\text{and}\; o$ are intermediate states of a token representation. $\sigma$ is SiLU activation and $\otimes$ is elementwise multiplication. \textit{Conv + SSM} operation abstracts the Conv1D and \textit{selective}-SSM operations.}
    \label{fig:mambablock}
\vspace{-25pt}
\end{wrapfigure}

Formally, Mamba is an autoregressive language model: $M: \mathcal{X} \rightarrow \mathcal{Y}$ over a vocabulary $\mathcal{V}$ that maps a sequence of tokens $x = [x_1, x_2, \dots, x_T] \in \mathcal{X},\; x_i \in \mathcal{V}$ to $y \in \mathcal{Y} \subset \mathbb{R}^{|\mathcal{V}|}$ which is a probability distribution over the next token continuations of $x$. Similar to other deep LMs, in Mamba, a token $x_i$ is first embedded to a hidden state of size $d$ as $h_i^{(0)} = emb(x_i)$. Then  $h_i^{(0)}$ is transformed sequentially by a series of MambaBlocks. The hidden state $h_i^{(\ell)}$ after the $\ell^{th}$ (1-indexed) MambaBlock is computed as follows:

\vspace{-10pt}
\begin{align}
h_i^{(\ell)} = h_i^{(\ell-1)} + o_i^{(\ell)}
\end{align}
\vspace{-15pt}

where $o_i^{(\ell)}$ is the output of $\ell^{th}$ MambaBlock for the $i^{th}$ token

\vspace{-10pt}
\begin{align}
    o_i^{(\ell)} & = \text{MambaBlock}^{(\ell)}\Big(h_1^{(\ell-1)}, h_2^{(\ell-1)}, \dots , h_{i}^{(\ell-1)}\Big)
         = \text{W}_{o}^{(\ell)}\,\Big(s_i^{(\ell)} \otimes g_i^{(\ell)}\Big)
    \label{eq:block_output}
\end{align}
\vspace{-15pt}

Here, $\otimes$ represents element-wise multiplication or Hadamard product. $s_i^{(\ell)}$ is calculated as:

\vspace{-15pt}
\begin{align}
    a_i^{(\ell)} &= \text{W}_a^{(\ell)} h_i^{(\ell)} \label{eqn:ssm-start}\\
    c_1^{(\ell)}, c_2^{(\ell)}, \dots, c_i^{(\ell)} &= \text{SiLU}\Big(\text{Conv1D}\Big(a_1^{(\ell)}, a_2^{(\ell)}, \dots, a_i^{(\ell)}\Big)\Big) \label{eqn:ssm-conv}\\
    s_i^{(\ell)} &= \text{\textit{selective}-SSM}\Big(c_1^{(\ell)}, c_2^{(\ell)}, \dots, c_i^{(\ell)}\Big)
    \label{eqn:ssm-end}
\end{align}
\vspace{-12pt}

We abstract the operations in Equations \ref{eqn:ssm-conv} and \ref{eqn:ssm-end} as the \textbf{Conv + SSM} operation in \Cref{fig:mambablock}. At a high level, Conv + SSM brings information from the past token representations to the current token representation. The purpose is similar to the \textit{attention} blocks in transformer LMs. But, unlike attention operation, Conv + SSM scales \textit{linearly} with the context length and thereby enjoys faster inference speed and longer context limits. See \cite{gu2023mamba} for details.

The output of the other path $g_i^{(\ell)}$ (that does not pass through Conv + SSM operation) is a gating mechanism that regulates the information flow.
This gating mechanism resemble parts of LSTM \citep{hochreiter1997long} and GRU \citep{cho2014properties} networks, where similar gates control selective updates of recurrent state.

\vspace{-15pt}
\begin{align}
g_i^{(\ell)} = \text{SiLU} \Big( \text{W}_g^{(\ell)} h_i^{(\ell-1)} \Big)
\end{align}
\vspace{-15pt}

In the remainder of the paper, we aim to characterize the role of the components of Mamba in factual recall by adapting tools that have previously been used to analyze transformers. In Section~\ref{sec:locating}, we apply activation patching to localize factual recall as in~\citet{meng2022locating}, testing the roles of states $s_i$, $g_i$, and $o_i$ at all layers. In Section~\ref{sec:editing}, following~\citet{meng2022locating,hase2024does}, we test rank-one edits of facts across components $\text{W}_a$, $\text{W}_g$, and $\text{W}_o$ at each layer. In Section~\ref{sec:lre}, we collect Jacobians within Mamba to test the linearity of relational encodings as done by~\citet{hernandez2023linearity}. And in Section~\ref{sec:attn_patch} we address the challenge of applying attention patching in Mamba, as used in~\citet{geva2023dissecting} to isolate information flow in GPT LMs.

\section{Locating Key States for Factual Recall}
\label{sec:locating}

\begin{figure*}
\vspace{-15pt}
    \centering
    \begin{subfigure}[t]{0.6\textwidth}
        \centering
        \includegraphics[width=.98\columnwidth]{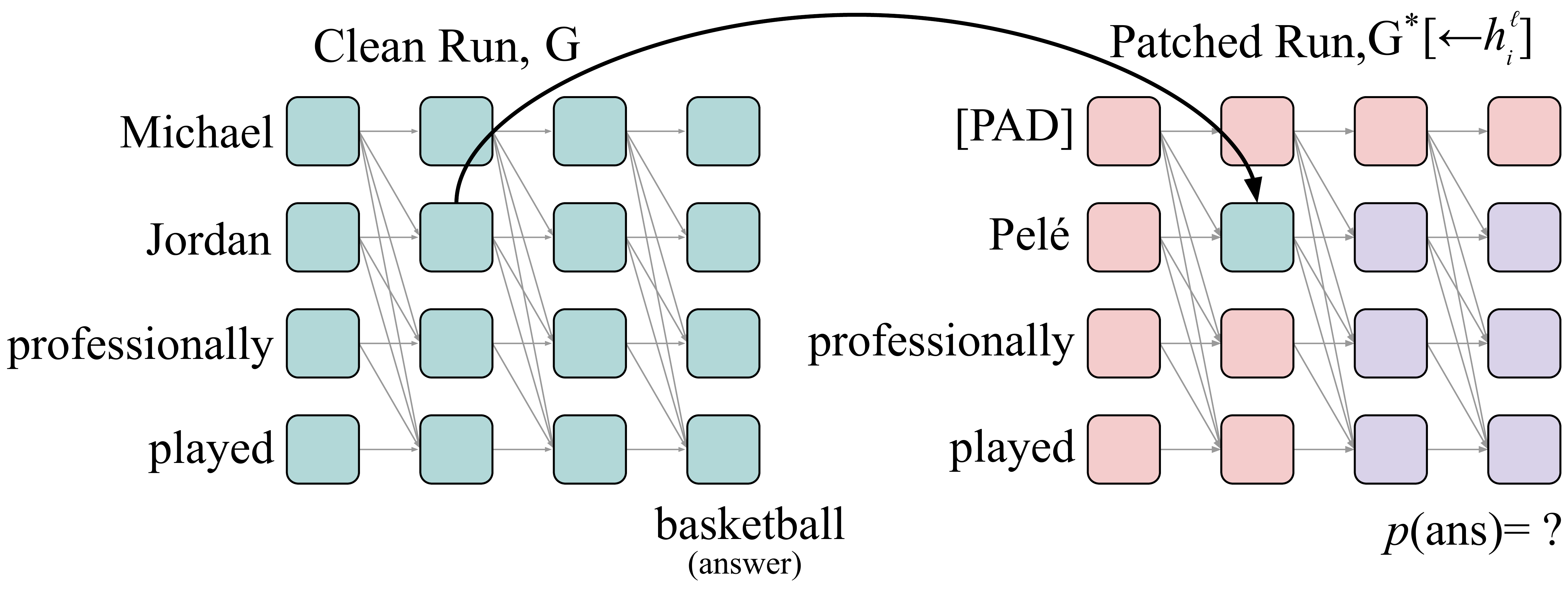}
        \caption{Activation patching}
        \label{fig:activation_patching}
    \end{subfigure}%
    \begin{subfigure}[t]{0.4\textwidth}
        \centering
        \includegraphics[width=\columnwidth]{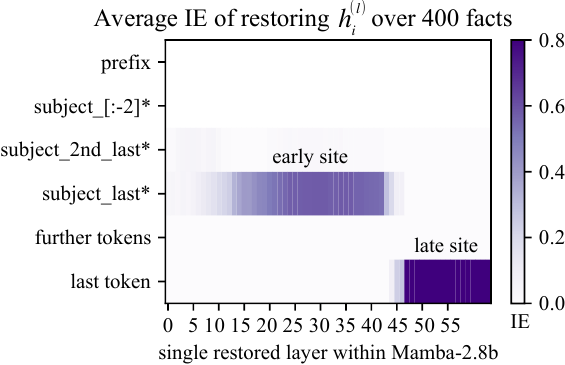}
        \caption{Tracing the residual states, $h_i^{(l)}$}
        \label{fig:trace_res}
    \end{subfigure}
    \vspace{-5pt}
    \caption{\small\textbf{(a)} 
    Activation patching. A state from the clean run $G$ is \textit{patched} into its corresponding position in the corrupted run $G^*$. This has a downstream effect of potentially changing all the states that depend on the patched state in $G^*[\leftarrow h_i^{(\ell)}]$. \textbf{(b)} Average indirect effect of applying causal tracing on residual stream states \big({$h_i^{(\ell)}$} in \Cref{fig:mambablock}\big) across 400 different facts from the \textsc{Relations} dataset (see \Cref{app:relations}).}
    \vspace{-15pt}
\end{figure*}

We begin with activation patching, seeking to understand if there are specific hidden states which play important roles during factual recall. We select a fact $(s, r, o)$ that the LM knows, where $r$ is a relation that associates a subject entity $s$ with an object entity $o$. To estimate each state's contribution towards a correct factual prediction $(s = \textit{Michael Jordan},\ r = \textit{professionally played},\ o = \textit{basketball})$, we collect model activations across three different runs:

\vspace{-5pt}
\paragraph{clean run} $G$: In the clean run, we simply run the model on a prompt specifying the fact we are interested in. For example, $x = (s, r) = \textit{Michael Jordan professionally played}$. We cache all the hidden states during the clean run to be used later: $\big\{h_i^{(\ell)}, a_i^{(\ell)}, s_i^{(\ell)}, g_i^{(\ell)} |\; i \in [1, T],\; \ell \in [1, L]\big\}$.

\vspace{-5pt}
\paragraph{corrupted run} $G^*$: In the corrupted run, we swap $s$ with a different subject $s^*(\textit{Pelé})$ such that the LM gives a different answer $o^*(\textit{soccer})$ to the modified prompt $x^*=(s^*, r)$ (i.e., $o^* \neq o$).

This subject-swapping approach follows the recommendation of \cite{zhang2023towards} and has the advantage of using natural text perturbations to avoid introducing out-of-domain states to the model's computation, as may happen when corrupting $s$ embeddings with Gaussian noise (the method used in \cite{meng2022locating}).

\vspace{-5pt}
\paragraph{patched run} $G^*[{\leftarrow h_i^{(\ell)}}]$:
In the patched run, we run the model on the corrupted prompt $x^*$, but intervene on $h_i^{(\ell)}$ by replacing its value with the corresponding state cached from the clean run $G$. The remainder of the computation is run normally, meaning that the patched state can have a downstream effect of potentially changing all the states that depend on it. See \Cref{fig:activation_patching}.

Let $p(o)$, $p^*(o)$, and $p^*[{\leftarrow h_i^{(\ell)}}](o)$ denote the probability assigned to the correct answer $o$ in $G$, $G^*$, and $G^*[{\leftarrow h_i^{(\ell)}}]$ respectively. To measure the contribution of $h_i^{(\ell)}$ in recalling the fact $(s, r, o)$, we define its \textit{indirect effect} (IE) as:
\begin{align}
\text{IE}_{h_i^{(\ell)}} = \frac{p^*[{\leftarrow h_i^{(\ell)}}](o) - p^*(o)}{p(o) - p^*(o)}
\end{align}

In \Cref{fig:trace_res} we plot the average indirect effect of restoring the residual states $h_i^{(\ell)}$ across different layer-token positions over 400 facts from the \textsc{Relations} dataset \citep{hernandez2023linearity}. The high IE observed at the \emph{late site} (later layers at the last token) position is natural, as restoring a clean $h_i^{(\ell)}$ there will restore most of the model computation from $G$. However, Mamba also shows high causality at the \emph{early site} (early-middle layers at the last subject token position). This is consistent with what \cite{meng2022locating} observed in the GPT family of language models.
\begin{figure}
\vspace{-15pt}
    \centering
    \includegraphics[width=\columnwidth]{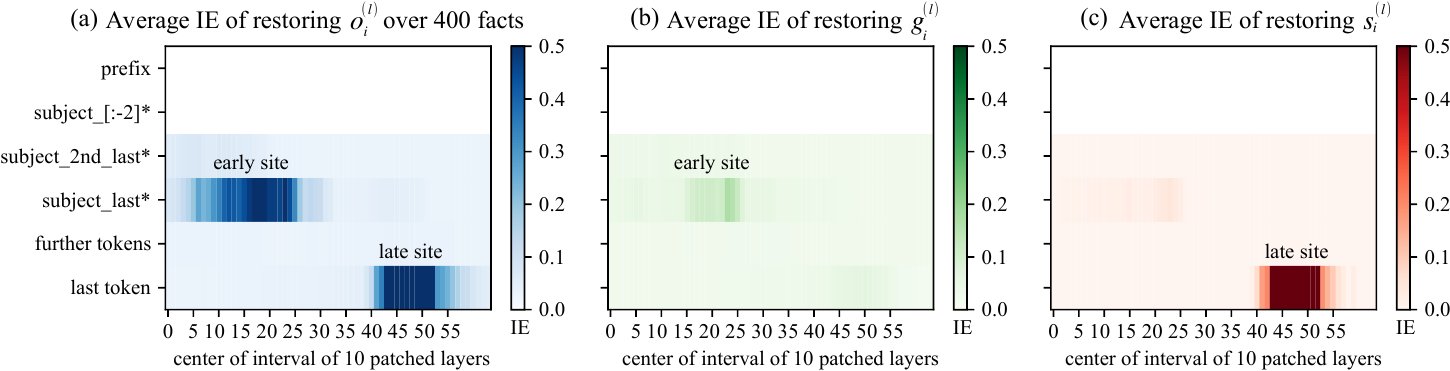}
    \caption{\small Average indirect effect of different states $o_i^{(\ell)}$, $g_i^{(\ell)}$, and $s_i^{(\ell)}$ over 400 facts from the \textsc{Relations} dataset (see \Cref{app:relations}). For each layer $\ell$, states for a window of 10 layers around $\ell$ are restored from the clean run $G$.}
    \label{fig:trace_state}
\vspace{-15pt}
\end{figure}%

\begin{wrapfigure}{R}{0.50\textwidth}
\vspace{-10pt}
    \centering
    \includegraphics[width=0.50\textwidth]{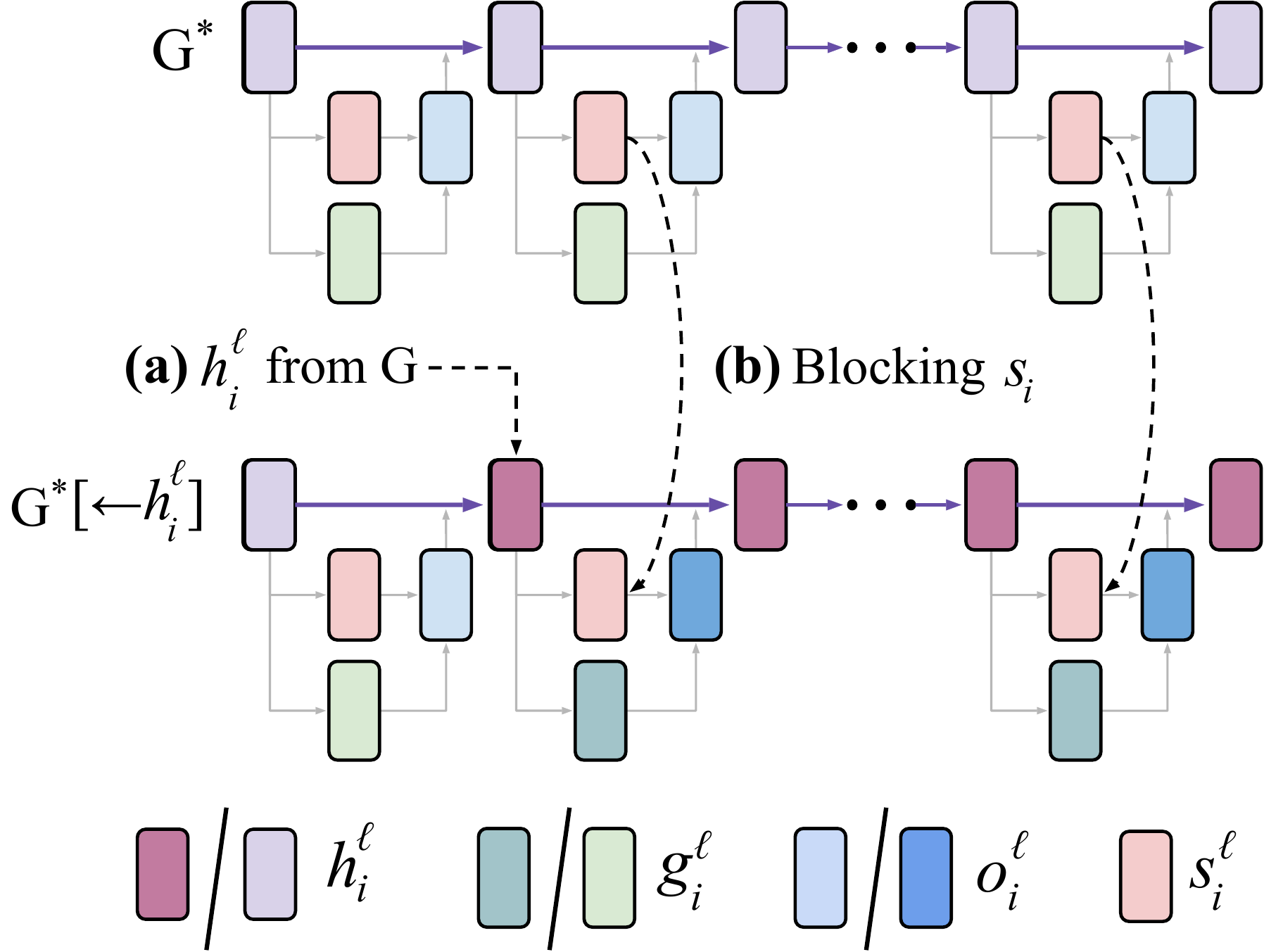}
    \vspace{-15pt}
    \caption{\small To probe for path-specific effects, \textbf{(a)} $h_i^{(\ell)}$ is restored from the clean run $G$ as in \Cref{fig:activation_patching}. \textbf{(b)} Then, to reveal the role of the Conv + SSM contributions, $s_i$ states from the corrupted run $G^*$ are also patched to block the contributions from those paths.}
    \vspace{15pt}
    \label{fig:path-blocking-setup}
\vspace{-35pt}
\end{wrapfigure}

In \Cref{fig:trace_state} we plot the average IE for $o_i^{(\ell)}$, $g_i^{(\ell)}$, and $s_i^{(\ell)}$. The plot for $o_i^{(\ell)}$ (\Cref{fig:trace_state}a)  looks very similar to \Cref{fig:trace_res}, confirming that the output from MambaBlock has strong causal effects at both early and late sites.  Interestingly, \Cref{fig:trace_state}c shows that the selective-SSM outputs $s_i^{(\ell)}$ have high IE only at the late site, resembling the behavior of attention modules in GPT models \citep{meng2022locating}. However, there is no state that appears to do the opposite; in other words, there is no state with strong effects at the early site and not at the late site (The gate output $g_i^{(\ell)}$ does have stronger IE at the early site, but these effects are very weak). 
To compare with autoregressive transformer LMs, activation patching results for Pythia-2.8b is shown in \Cref{fig:trace_pythia} in \Cref{app:locate-pythia}.
This comparison reveals a key way how Mamba differs from transformers: while transformer MLP outputs have effects in the early site and not the late site, in Mamba there is no similar state that specializes only at the early site, at which factual recall would be expected to occur. This presents the question: which parameters in Mamba mediate factual recall?

To investigate this question, we replicate an experiment from \cite{meng2022locating} to probe \textit{path-specific effects} \citep{pearl2022direct} by severing a path from the causal graph and monitoring its effect. Here, we are interested in understanding the effect of the contributions from $g_i$, $s_i$, and $o_i$ (i.e. states that are processed by $\text{W}_g$, Conv + SSM, and $\text{W}_o$ respectively) while recalling a fact. First, in the corrupted run $G^*$, at token position $i$, we cache all the contributions from the $s_i$ paths as $s_i^* = \{s_i^{*(\ell)} |\; \ell \in [1, L]\}$. Then in the patched run $G^*[{\leftarrow h_i^{(\ell)}}]$, we restore $h_i^{(\ell)}$ that was cached from the clean run $G$ into its corresponding state (as in \Cref{fig:activation_patching}), but with an additional modification: to understand the contribution from the $s_i$ paths, we sever those paths by also patching $s_i^*$ (cached from the corrupted run $G^*$) to their corresponding locations (see \Cref{fig:path-blocking-setup}). The same experiment is replicated to understand the contributions of $g_i$ and $o_i$ states. We note that severing the $o_i^{(\ell)}$ will sever $s_i^{(\ell)}$ and $g_i^{(\ell)}$ as well (see \Cref{fig:mambablock}). 
\begin{figure}
\vspace{-15pt}
    \centering
    \includegraphics[width=.95\columnwidth]{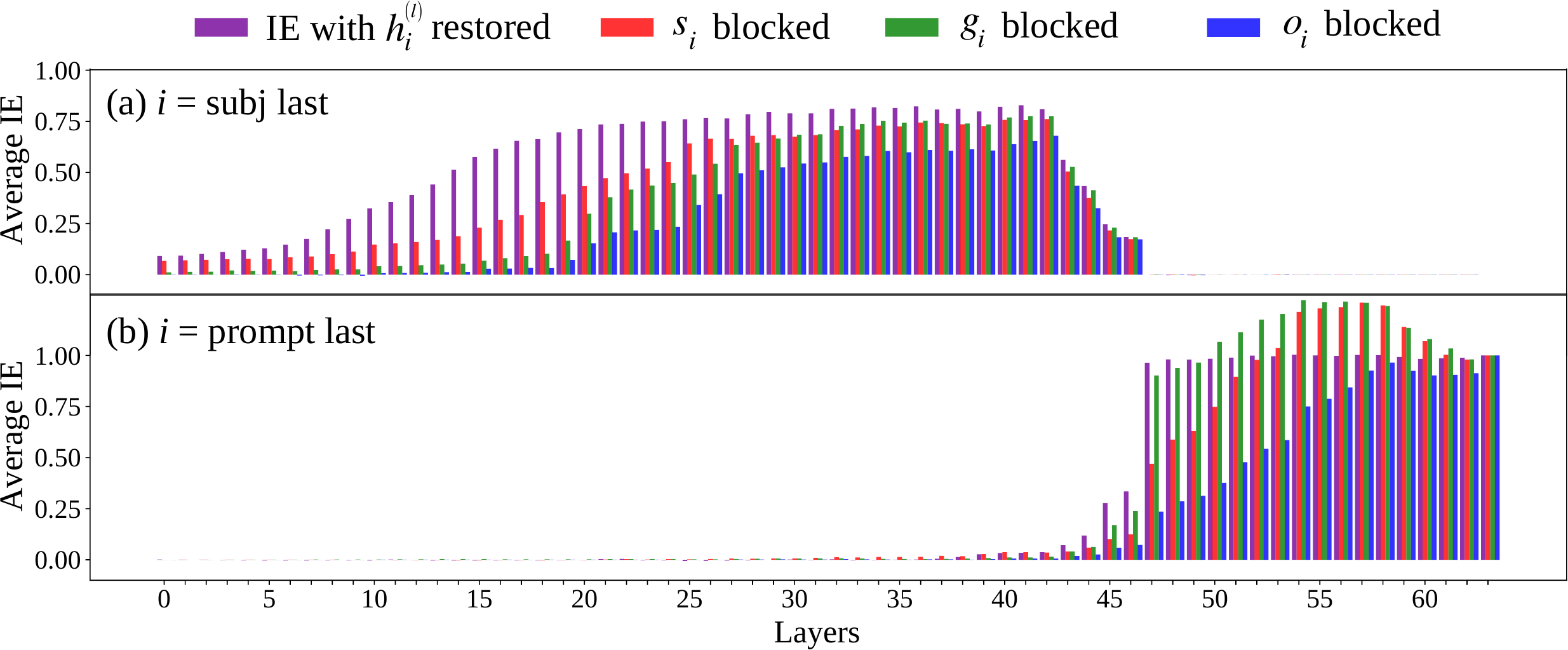}
    \caption{\small Impact of ablating $s_i$, $g_i$, and $o_i$ on $\text{IE}_{h_i^{(\ell)}}$ for \textbf{(a)} \textit{subject last}  and \textbf{(b)} \textit{prompt last} token positions. Taken together (a) and (b) show a clear separation roles between early-mid and later layers in Mamba-2.8b. $h_i^{(\ell)}$ up to layer $46$ only show strong IE at the subject last token position and have negligible impact after that. Whereas IE of $h_i^{(\ell)}$ jumps to $1.0$ after layer $46$. \textbf{(a)} also shows that, at the subject last token, before layer $27-28$, $\text{IE}_{h_i^{(\ell)}}$ is significantly reduced by blocking either $o_i$, $g_i$, or $s_i$ paths \big(sorted in descending order of damaging $\text{IE}_{h_i^{(\ell)}}$\big). \textbf{(b)} At the prompt last token, ablating $o_i$ or $s_i$ paths can significantly reduce $\text{IE}_{h_i^{(\ell)}}$ in layers $47-50$.  
    }
    \label{fig:sub_last_path}
\vspace{-16pt}
\end{figure}

In \Cref{fig:sub_last_path} we plot the average results of this experiment for token positions (a) $i = \textit{subject last}$ and (b) $i = \textit{prompt last}$ over 400 examples randomly sampled from the \textsc{Relations} dataset. The key findings can be understood by examining the gap between the purple bars and the green, red, and blue bars: a large gap indicates a strong mediating role for Conv + SSM, $\text{W}_g$, or $\text{W}_o$ parameters, respectively.  At the early site at the subject last token, both the Conv + SSM and $\text{W}_g$ have a strong role, but $\text{W}_o$ plays an even larger role than either.  Yet the strongest mediator at the late site is also $\text{W}_o$. This experiment highlights the importance of $\text{W}_o$ in both stages of predicting a fact. But it also suggests that Mamba does not separate early-site factual recall between these groups of parameters as cleanly as transformers. However, \Cref{fig:sub_last_path} reveals a clean separation of roles between early to mid and later layers, analogous to the findings of \cite{hernandez2023linearity} in transformer LMs. We also note that this division of responsibilities between layers can be more sharply noticed in Mamba when compared to transformers LMs (compare \Cref{fig:sub_last_path} with \Cref{fig:path_blocking_pythia}).

\section{Editing Facts With ROME}
\label{sec:editing}   

\label{sec:rome}
Having begun to characterize the locations of important states for factual recall, we now investigate whether factual recall behavior can be edited. In particular, we apply the ROME~\citep[Rank One Model Editing,][]{meng2022locating} technique to Mamba. ROME begins with the observation that any linear transformation can be considered as an associative memory \citep{anderson1972simple, kohonen1972correlation}, mapping a set of keys $K = [k_1 | k_2 | \dots]$ to their corresponding values $V=[v_1 | v_2 | \dots]$, and uses this to edit factual associations in transformer LMs. Here, we apply the technique to a particular set of linear transformations within Mamba, and report our editing success on each.\footnote{Further motivating these experiments, previous work has shown that the locations identified by activation patching techniques are not necessarily those which have the strongest edit performance ~\citep{hase2024does}.}%

The input to ROME is a prompt $x = (s, r)$, where $s$ (\textit{Emmanuel Macron}) is a subject entity and $r$ (\textit{is the President of}) is a relation. ROME also takes a counterfactual object $o^*$ (\textit{England}), meant to replace the correct object $o$ (\textit{France}) in the model's output. To effect that change, ROME generates a rank-one update to $\text{W}_\mathit{down}^{(\ell)}$, the down-projection matrix of the MLP module for the last token of the subject at layer $\ell$---which plays the role of the associative memory. In generating the rank-one update, ROME considers the input to $\text{W}_\mathit{down}^{(\ell)}$ as the \textit{key} ($k_*$). Then, with gradient descent ROME calculates a \textit{value} ($v_*$) such that, when $v_*$ is inserted as the output of $\text{W}_\mathit{down}^{(\ell)}$, the model will output $o^*$. Importantly, while optimizing $v_*$, ROME attempts to minimize unrelated changes in model outputs (\textit{Joe Biden}, for example, should still be mapped to \textit{the United States} post-edit). Finally, ROME adds a rank-1 matrix $\Delta$ to $\text{W}_\mathit{down}^{(\ell)}$ such that $\big(\text{W}_\mathit{down}^{(\ell)} + \Delta\big)k_* \approx v_*$. (See \cite{meng2022locating} for details.)

\subsection{Applying ROME in Mamba}
\label{sec:applying_rome}

\begin{figure}
\vspace{-15pt}
    \centering
\includegraphics[width=\columnwidth]{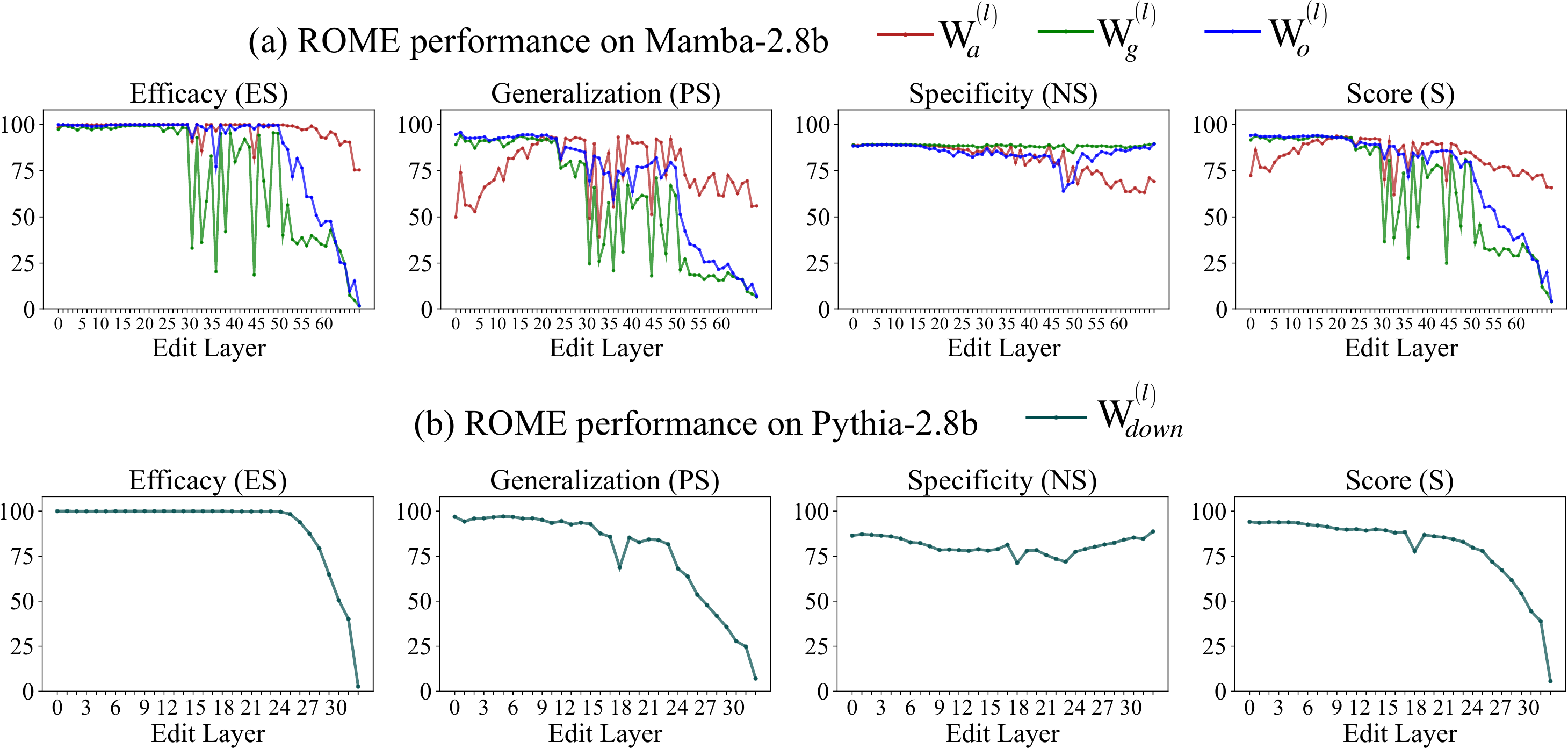}
    \caption{
    ROME performance in editing facts across different layers \textbf{(a)} by modifying $\text{W}_a^{(\ell)}$, $\text{W}_g^{(\ell)}$, $\text{W}_o^{(\ell)}$ in Mamba-2.8b, and \textbf{(b)} modifying $\text{W}_{down}^{(\ell)}$ in Pythia-2.8b. Results are reported on the first 2000 examples in the \textsc{CounterFact} dataset.
    }
    \label{fig:rome_performance}
    \vspace{-10pt}
\end{figure}

We apply ROME on the three different projection matrices of Mamba: $\text{W}_a^{(\ell)}$ which affects only the Conv + SSM path, $\text{W}_g^{(\ell)}$ which affects only the gating path, and $\text{W}_{o}^{(\ell)}$, the final output of the MambaBlock, which is added to the residual state. We plot ROME performance on different projection matrices \big($\text{W}_a^{(\ell)}$, $\text{W}_g^{(\ell)}$, and $\text{W}_o^{(\ell)}$\big) across all the layers in \Cref{fig:rome_performance}a.

To evaluate editing performance, we use the \textsc{CounterFact} dataset from \cite{meng2022locating}. \textsc{CounterFact} contains 20K counterfactual examples in the form $(s, r, o \rightarrow o^*)$, where $o$ is the correct answer to the prompt $x=(s, r)$, and $o^*$ is the object which is to be inserted as the new answer to the prompt (See \Cref{app:counterfact} for details). We select the first 2000 examples from this dataset for our module-layer sweep. We use the original evaluation matrices in \cite{meng2022locating} to measure ROME edit performance.

The final \textbf{score (S)} in the ROME evaluation suite is the harmonic mean of three different scores:

\vspace{-5pt}
\begin{enumerate}[leftmargin=5.5mm]
\itemsep -0.5pt 

    \item \textbf{Efficacy (ES)}: For an edit request $(s, r, o \rightarrow o^*)$, we say the edit is \textit{effective} if, post-edit, the LM assigns $p(o^*) > p(o)$ in response to the prompt $x = (s, r)$. Efficacy reflects the portion of the examples where the edit was effective.
    
    \item \textbf{Generalization (PS)}: A successful edit should be persistent across different paraphrases of $(s,r)$.  For each of the request instances $(s, r, o \rightarrow o^*)$, $p(o^*) > p(o)$ is checked post-edit with a set of different rephrasings $x_p \sim \mathcal{P}_r(s)$ of the prompt  $x = (s, r)$, where $\mathcal{P}_r$ denotes a set of paraphrased templates for the relation $r$.
    
    \item \textbf{Specificity (NS)}: Finally, the edit should be specific to $\mathcal{P}_r(s)$ and should not additionally change the mapping of some nearby subject $s_n$ to $o^*$. To evaluate the specificity of an edit we measure $p(o_n) > p(o^*)$ with $\mathcal{P}_r(s_n)$ for a set of nearby factual associations $\{(s_n, r, o_n) \,|\, o_n \neq o^*\}$.
    
\end{enumerate}

\Cref{fig:rome_performance}a shows that ROME can achieve high scores (S) for a range of early to middle layers by modifying any one of the projection matrices $\text{W}_a^{(\ell)}$, $\text{W}_g^{(\ell)}$, or $\text{W}_{o}^{(\ell)}$, matching observations made by \cite{hase2024does} regarding transformer LMs.
However, we found that performance does depend on the location of the edit. For example, in the case of $\text{W}_g^{(\ell)}$ and $\text{W}_o^{(\ell)}$, the score (S) and generalization (PS) drops after around layer 43. This is consistent with our findings from the path-blocking experiment in \Cref{fig:sub_last_path}a. We also find that edits to $\text{W}_a^{(\ell)}$ have poor generalization (PS) in early layers, whereas high PS can be achieved at early layers by modifying either $\text{W}_g^{(\ell)}$ or $\text{W}_o^{(\ell)}$, consistent with their higher indirect effects as seen in \Cref{fig:sub_last_path}a.

Where is the right place to apply ROME on Mamba? \Cref{fig:trace_state} could suggest $\text{W}_g^{(\ell)}$, since the causal effect of $g_i$ states is mostly concentrated at the subject last token, similar to the behavior of MLPs in transformers \citep{meng2022locating}. Consistent with this is the architectural fact that, just as transformers' $\text{W}_{down}^{(\ell)}$ connects to attention modules only through the residual stream, the output of $\text{W}_g^{(\ell)}$ does not flow through the Conv + SSM module---a module that other work has suggested might play a role similar to that played by attention heads in transformers \citep{grazzi2024capable}. And, indeed, we find that ROME can successfully insert facts by modifying $\text{W}_g^{(\ell)}$. On the other hand \Cref{fig:rome_performance}a reveals sudden drops in efficacy and generalization at middle layer gates, suggesting that $\text{W}_g^{(\ell)}$ may be an unreliable mediator at some layers. Our experiments further show that the best performance for ROME is empirically achieved by modifying $\text{W}_o^{(\ell)}$. This is consistent with the fact that $o_i$ states show a stronger causal effect at the subject last token than $g_i$ states do (see Figures \ref{fig:trace_state}a and \ref{fig:sub_last_path}a). Additionally, ROME achieves better generalization (PS), competitive specificity (NS), and an overall better score (S) with $\text{W}_o^{(\ell)}$. We hypothesize that the strong performance of $\text{W}_o^{(\ell)}$ may be due to the the separation of roles between early-mid and later layers observed in Figures \ref{fig:trace_res}, \ref{fig:trace_state}a, and \ref{fig:sub_last_path}. Also see \Cref{app:subtract} where we isolate the contribution of $\text{W}_o^{(\ell)}$ by subtracting $\text{IE}_{s_i^{(\ell)}} + \text{IE}_{g_i^{(\ell)}}$ from $\text{IE}_{o_i^{(\ell)}}$, which reveal a critical role of $\text{W}_o^{(\ell)}$ in early-mid layers at subject last token position while mediating a fact.

We plot ROME performance for a similar sized Pythia model on \Cref{fig:rome_performance}b  for comparison.

\section{\textsc{Linearity of Relation Embedding (\lre)}}
\label{sec:lre}

With activation patching we can identify \textit{where} facts are located inside a LM. We are also interested in understanding \textit{how} LMs extract this information given $x = (s, r)$. 
Figures \ref{fig:trace_res} and \ref{fig:sub_last_path} show a clear separation of roles in early-mid and later layers in Mamba. We observe a similar phenomenon in autoregressive transformer LMs \citep{meng2022locating, meng2022mass,  geva2023dissecting}. According to \cite{geva2023dissecting}, in transformer LMs, the subject entity representation $\mathbf{s}$, at the subject last token position, goes through an \textit{enrichment} process, mediated by the MLP in the early-mid layers, where $\mathbf{s}$ is populated with different facts/attributes relevant to the subject entity $s$. Then, at the last token position, attention modules perform a \textit{query} on the \textit{enriched} $\mathbf{s}$ to extract the answer to the prompt $x = (s, r)$. \citet{hernandez2023linearity} approximate the \textit{query} operation performed on the \textit{enriched} $\mathbf{s}$ for a specific relation $r$ by taking the first order Taylor series approximation (\textsc{Lre}) of the LM computation $F$ as

\vspace{-15pt}
\begin{align}
    \label{eq:order_1_approx}
    F(\subjrep, r) & \approx \beta \, \text{J}_{\rho} \subjrep + b \nonumber   \\ 
    \text{where } \text{J} = \mathbb{E}_{\subjrep_i, r}\left[\left.\frac{\partial F}{\partial\subjrep}\right|_{(\subjrep_i, r)} \right] & \text{\;, \;\;\;}
    b = \mathbb{E}_{\subjrep_i, r}\left[\left.F(\subjrep, r) - \frac{\partial F}{\partial\subjrep} \; \subjrep\right |_{(\subjrep_i, r)} \right]  \text{\;, \;\;\;}\\
    \beta \; \text{is a scalar} & \text{\;, and \;\;\;} \rho \; \text{is the rank of $\text{J}$} \nonumber 
\end{align}
\vspace{-15pt}

\citet{hernandez2023linearity} show that for a range of different relations it is possible to achieve a \textsc{Lre} that is faithful to the model computation $F$ by averaging the approximations of $\text{J}$ and $b$ calculated on just $n=5$ examples. We utilize \textsc{Lre} to understand the complexity of decoding \textit{factual} relations in Mamba. We find the hyperparameters $\beta$, $\rho$ and the layer $\ell$ (where to extract the enriched $\mathbf{s}$ from) using grid search. For mathematical and implementation details, see \citet{hernandez2023linearity}.  

\begin{figure}
    \centering
    \includegraphics[width=.8\columnwidth]{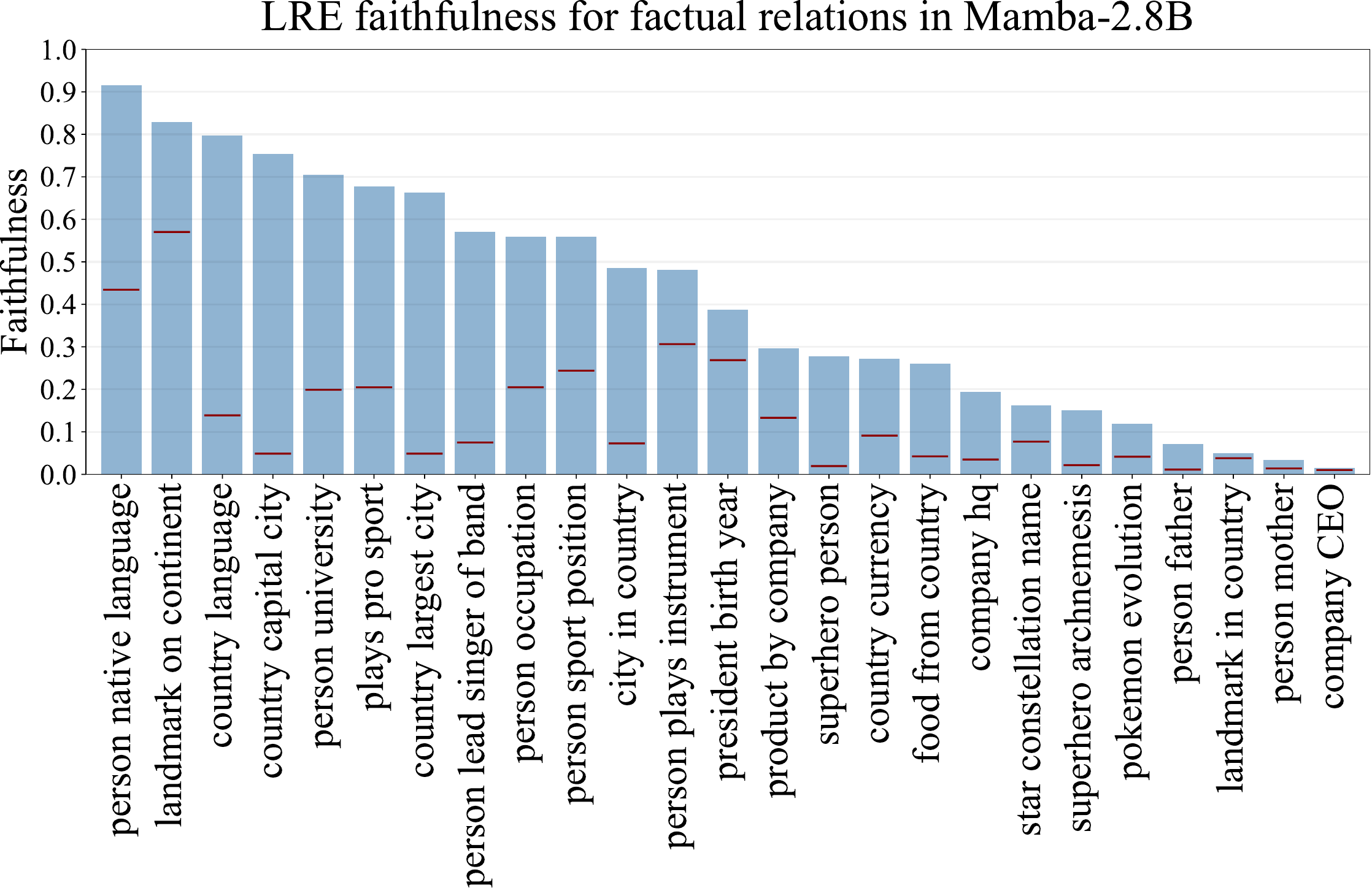}
    \vspace{-10pt}
    \caption{\small \textsc{Lre} \textit{faithfulness} with $n=5$ samples for all the factual relations. Horizontal red lines indicate random choice baseline (in the \textsc{Relations} dataset).}
    \label{fig:lre_mamba_faith}
    \vspace{-10pt}
\end{figure}

We plot the \textit{faithfulness} of \textsc{Lre} with $n=5$ samples on \Cref{fig:lre_mamba_faith}. The metric \textit{faithfulness} represents the portion of facts $(s, r, o)$ that can be correctly retrieved if the LM computation $F(\subjrep, r)$ is replaced with $\textsc{Lre}(\mathbf{s})$, a simple affine transformation. 

We only calculate $\textsc{Lre}$ for the \textit{factual} relations in the \textsc{Relations} dataset. \Cref{fig:lre_mamba_faith} shows that only for $10$ out of $26$ factual relations can a \textit{linear} \textsc{Lre} achieve more than $50\%$ \textit{faithfulness}. For comparison, in the same sized Pythia-2.8b \textsc{Lre} achives $>50\%$ \textit{faithfulness} for $11$ factual relations (see \Cref{app:lre-pythia}). And, in both Mamba and Pythia, \textsc{Lre} fails to achieve good \textit{faithfulness} for the relations where the \textit{range} (the number of unique answers) is large. These findings align with what \cite{hernandez2023linearity} observed on GPT and LLaMA models; suggesting that, similar to transformer LMs, factual knowledge might be heterogeneously represented for different relations in Mamba.

\section{Attention Knock-out in Mamba?}
\label{sec:attn_patch}

Attention modules mediate the flow of information across different token positions in transformer LMs. In attention \emph{``knock-out''} experiments the information that flows through a specific edge (from $k^{th}$ token to $q^{th}$ token) via a certain attention head is blocked to understand if critical information flows through that edge. This is also a form of causal mediation analysis and it has been effective in understanding the information flow in transformer LMs \citep{geva2023dissecting, wang2022interpretability, todd2023function}. In Mamba, information from past tokens is retained in the $s_i$ states, with the Conv + SSM operations (see \Cref{fig:mambablock} and Equations \ref{eqn:ssm-start}--\ref{eqn:ssm-end}). We ask, can we perform experiments similar to attention knock-out experiments in Mamba in order to understand how it moves factual information?

\textbf{We find that performing similar experiments in Mamba can be difficult}. The use of Conv with a non-linearity in conjunction with selective-SSM make it challenging to remove the information retained in the $q^{th}$ token from the $k^{th}$ token (see \Cref{app:ablate-retention} for details). However, it is possible to block the propagation of information from the $k^{th}$ token to \textit{all} the future tokens via Conv + SSM operation by mean-ablation. Specifically, for a layer $\ell$, we set $a_k^{(\ell)} := \mathbb{E}\big[a^{(\ell)}\big]$, where $\mathbb{E}\big[a^{(\ell)}\big]$ is the mean of $a^{(\ell)}$ states collected with 10,000 tokens from WikiText-103 by \cite{merity2016pointer}. We recognize that this intervention may not be as \textit{surgical} as cutting a specific edge. However, with some caveats, this experiment suggests that the factual information flow in Mamba is similar to what \cite{geva2023dissecting} observed in GPT LMs.

We randomly sample 700 facts across 6 factual relations from the \textsc{Relations} dataset. 
For each of those examples we block-out information propagation of the \textit{subject}, \textit{non-subject}, and the \textit{prompt-last} token positions for a window of $10$ layers around a specific layer $\ell$. The effect of blocking out $\text{Conv + SSM}$ information flow for certain layer-token ($\ell-k$) positions is measured as the relative change in $p(o)$ with $\nicefrac{\Big(p\big(o \,|\, a_k^{(\ell)} := \mathbb{E}\big[a^{(\ell)}\big]\big) - p(o)\Big)}{p(o)}$. \Cref{fig:attn_knockout} shows the averaged result and it leads us to draw the following conclusions about how factual information flows in Mamba:

\begin{figure}[h!]
    \centering
    \includegraphics[width=0.8\columnwidth]{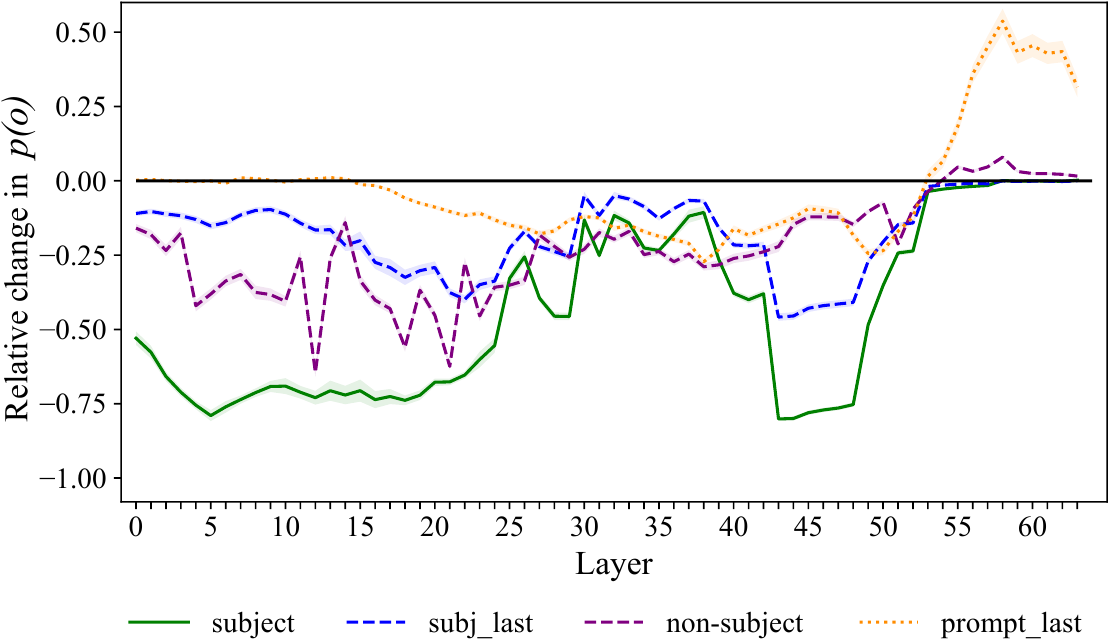}
    \caption{\small Relative change in $p(o)$ when information flow from $a_k^{(\ell)}$ to future tokens via $s_i$ paths is blocked, with $k$ taking the value of either \textit{subject}, \textit{non-subject}, or the \textit{prompt\_last} token positions. For each layer $\ell$, $s_i$ paths were blocked for a window of 10 layers around $\ell$.}
    \label{fig:attn_knockout}
\vspace{-0.5em}
\end{figure}

\begin{itemize}[leftmargin=5.5mm]
\itemsep -0.1pt

    \item[\textbf{(a)}] The purple lines show that blocking out non-subject information flow in early middle layers can bring down $p(o)$ by up to $50\%$. Non-subject tokens are used to specify the relation $r$. This observation leads us to believe that Mamba propagates relation specific information to future tokens using Conv+SSM operations in early-middle layers.

    \item[\textbf{(b)}] Interestingly, the green lines (blocking the subject information flow) shows two valleys:
    
    \begin{enumerate} [leftmargin=5.5mm]
    \itemsep -0.1pt
    
        \item The first valley at the early layers is not surprising as Mamba needs to collate information from all the subject tokens in early layers to recognize a subject entity $s$ consisting of multiple tokens.
        
        \item However, the valley at layers 43-48 suggest that Mamba uses Conv + SSM paths in those layers to propagate critical information from the subject to later tokens. This aligns with Figures \ref{fig:sub_last_path}b and \ref{fig:trace_state}c, where $s_i$ states in those layers show high indirect effects, indicating their crucial role while recalling a fact.
    
    \end{enumerate}

    \item[\textbf{(c)}] The blue dashed lines indicate the effect of blocking the information of only the subject last token. If the ablation is performed in very early layers, later layers can start to compensate for that. However, the valley around layers 20-21 suggests that Mamba expects to recognize the full subject entity by then in order to recall relevant associations (\textit{enrichment}). Notably, activation patching results for $o_i$ and $s_i$---states that we hypothesize take crucial part in the enrichment process---also show strong indirect effect around that region (Figures \ref{fig:trace_state}a, \ref{fig:trace_state}b, and \ref{fig:sub_last_path}a). The blue line follows the green line after layer 30. The weaker effect observed might be because ablating subject last token is not always enough to remove all the subject information. For example, in \textit{Eiffel Tower}, \textit{Eiffel} (tokenized as \texttt{E}, \texttt{iff}, \texttt{el})  is more informative than the last token \texttt{Tower}. 
\end{itemize}

These findings align with how factual information flows through attention modules in autoregressive transformer LMs, as observed by \cite{geva2023dissecting} in GPT. However, unlike \cite{geva2023dissecting}, we cannot make strong claims about the unique role of the final token position (prompt-last) with this experiment. As we block out information flow to \textit{all} future tokens, the intermediate states in between the ablated $k^{th}$ token and the last token are affected as well.  

\section{Related Works}
\label{sec:related}   

\vspace{-5pt}
\paragraph{Mamba.} Mamba is a recent family of language models that are based on state space models (SSMs). Neural SSM-based models have achieved good performance across different modalities, including vision~\citep{nguyen2022S4ND}, audio~\citep{pmlr-v162-goel22a}, and genomic sequences~\citep{nguyen2023hyenadna}. Only recently, however, with Mamba, have they become competitive with the language modeling performance of transformers~\citep{gu2023mamba}. Like transformers, Mamba contains factual knowledge about real world entities~\citep{grazzi2024capable}. However, knowledge representation in Mamba (and other LMs based on SSMs) has up to now remained unexplored.
 
There are few works focused on interpreting Mamba. \cite{ali2024hidden} identify implicit attention-like matrices formed by Mamba's selective state space layers. \cite{grazzi2024capable}, while not strictly focused on interpreting Mamba's internals, apply linear probes to Mamba's (decoded) intermediate states during in-context regression tasks. Like us, they find substantial similarities between Mamba and transformer models: both architectures pursue ``iterative'' strategies, with the task loss falling more or less monotonically as the layer index increases.

\noindent
\textbf{Locating Factual Knowledge in Language Models.} To make factually correct statements about the world, a LM has to store factual knowledge about real world entities  somewhere in its parameters. Understanding how and where a neural network stores knowledge is a core problem for interpretability and it has thus been studied from several perspectives \citep{ji2021survey, wang2014knowledge}. One line of work trains classifiers to probe for properties encoded in model representations \citep[]{ettinger2016probing,shi2016does,hupkes2018visualisation,conneau2018you,belinkov2017neural,belinkov2019analysis}. However, the flexibility of these classifiers can lead to overestimating model knowledge and capabilites~\citep{belinkov2022probing}. Causal mediation analysis methods \citep{pearl2022direct} attempt to measure the causal contribution of intermediate states to task performance. \cite{meng2022locating, meng2022mass} use activation patching to identify key MLP modules for factual recall, highlighting the middle layers at particular token positions as being especially important. Similarly, \cite{geva2023dissecting} apply causal mediation analysis to attention modules, seeking to understand the mechanism of cross-token factual information flow inside transformer LMs.

\section{Discussion}
\label{sec:conclusion}  

\vspace{-.3em}
In this paper we have set out to understand whether the analytical methods and tools developed for transformer LMs can also be applied on the Mamba recurrent state-space architecture.
Although our experiments have been limited to Mamba-2.8b, the largest available LM of that family, and comparisons to the similarly-sized transformer Pythia-2.8b,
the methods we have introduced are general, and can be used to analyze to any state-space model.

Our overall comparisons of Mamba and transformers are positive: with activation patching we have found that, similar to autoregressive transformer LMs, Mamba shows signs of localization at the last subject token and at specific layer ranges while recalling a fact.
Although, unlike transformers, Mamba has no MLP modules, we find that their $\text{W}_o$ weights can receive rank-one model editing (ROME) edits with good generalization and specificity at a range of layers, similar to $\text{W}_\mathit{down}$ in Pythia and GPT family of LMs. We have studied the linearity of the embeddings of factual relations in Mamba and have found that many can be well approximated by $\textsc{Lre}$, again resembling autoregressive transformer LMs.
We have also been able to partially adapt the tools of attention knock-out in Mamba by blocking outgoing information from a token, revealing information flows similar to transformer LMs during factual recall.

The similarity that we have observed between factual recall mechanisms in transformers and Mamba leads us to speculate that the autoregressive language modeling task itself induces a pattern of localized factual recall that is independent of modeling architecture. When constraining a model to process text from beginning to end, the ordering creates a specific bottleneck in the information flows: the end of a subject becomes a singular moment at which recognition of the subject is both possible and useful, and we find that both transformers and Mamba arrange their computations to localize factual recall at that moment. We hypothesize that other future autoregressive LMs architectures should expect to see similar locality in factual recall as well.

In summary, we find that many of the tools used to interpret and edit large transformers can be adapted to work with Mamba, and we are optimistic that those tools will continue to be useful as architectures continute to evolve.

\section*{Ethics}
\label{sec:ethics}   

By exploring the factual recall mechanism in Mamba, we potentially improve its transparency, enabling oversight and control. However, the ability to modify facts directly in the model brings with it the potential for abuse, such as adding malicious misinformation or bias.
\section*{Reproducibility}
\label{sec:repro}   

We ran all experiments on workstations with either
80GB NVIDIA A100 GPUs or 48GB A6000 GPUs, using the HuggingFace Transformers library \citep{wolf2019huggingface} and PyTorch \citep{paszke2019pytorch}. We make use of publicly available datasets \textsc{CounterFact} and \textsc{Relations} in this work. 
\section*{Acknowledgements}
\label{sec:ack}   

This research has been supported by a grant from Open Philanthropy (DB, AS), and an NSF Computer and Information Science and Engineering Graduate Fellowship Fellowship (DA). We are also grateful to the Center for AI Safety (CAIS) for sharing their compute resources, which supported many of our experiments. Some of our initial analyses were conducted with a beta version of \texttt{NNsight}~\citep{fiotto2024nnsight} on an implementation of Mamba instrumented for research by Jaden Fiotto-Kaufmann.

\bibliography{references}
\bibliographystyle{colm2024_conference}

\newpage
\appendix

\section{Datasets}
\label{app:dataset}
We use two datasets; \textsc{CounterFact} by \cite{meng2022locating} and \textsc{Relations} by \cite{hernandez2023linearity} in this work.

\subsection{\textsc{CounterFact}}
\label{app:counterfact}
\cite{meng2022locating} developed the \textsc{CounterFact} dataset for evaluating the efficacy of counterfactual edits in language models. It was prepared by adapting \textsc{ParaRel} (\cite{elazar2021measuring}) and scraping \textit{Wikidata}\footnote{\href{https://www.wikidata.org/wiki/Wikidata:Main_Page}{\texttt{www.wikidata.org/wiki/Wikidata:Main\_Page}}}. The dataset contains $21,919$ requests $\{s, r, o, o^*, \pi^*\}$ where $o$ is the correct answer to the prompt $x=(s, r)$, $o^*$ is the counterfactual edit request, and $\pi^* \sim \mathcal{P}(s, r)$ is a paraphrase of the prompt $x=(s, r)$ to test for generalizability (PS). Each of the records also contain some neighborhood prompts $\pi_N$ to test for specificity (NS) and some generation prompts $\pi_G$ to test if LM generation post-edit is fluent and consistent with the edit. Please refer to \cite{meng2022locating} for details on the curation of this dataset.

We evaluate ROME performance in Mamba-2.8b (\Cref{fig:rome_performance}a) and Pythia-2.8b (\Cref{fig:rome_performance}b) on the first $2000$ records from \textsc{CounterFact}. 

\subsection{\textsc{Relations}}
\label{app:relations}
The \textsc{Relations} dataset introduced in \cite{hernandez2023linearity} consists of $47$ relations of $4$ types: \textit{factual}, \textit{linguistic}, \textit{bias}, and \textit{commonsense}. A relation $r$ is an association between two entities. For example, the relation, $r = \textit{professionally played the sport}$ connects the subject $s = \textit{Michael Jordan}$ with the object $o = \textit{basketball}$. The dataset contains a set of $(s, o)$ for each relation $r$.

\begin{figure}[h!]
    \centering
    \includegraphics[width=\columnwidth]{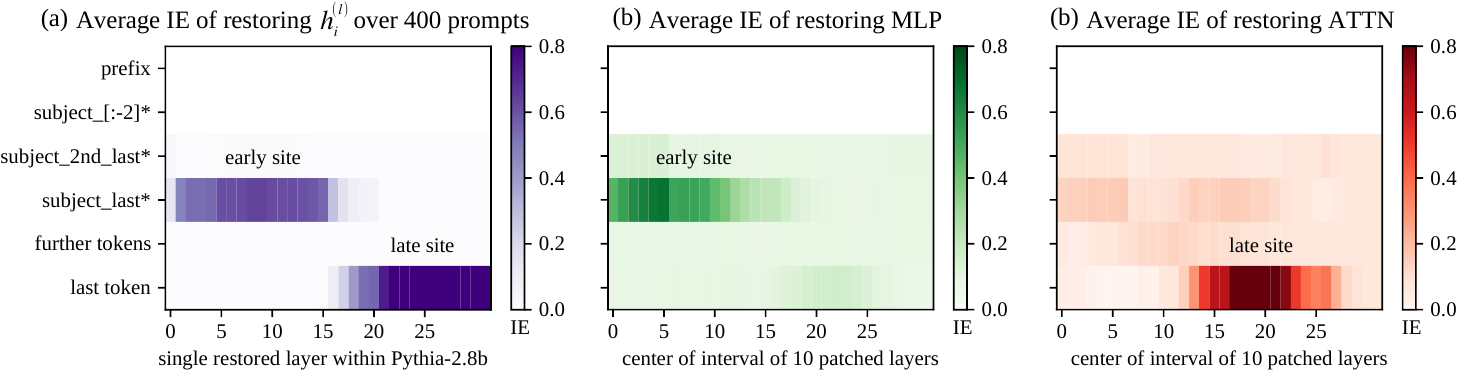}
    \caption{Average indirect effect of residual state, MLP, and attention outputs in Pythia-2.8b over 400 facts. For MLP and attention outputs a window of 10 layers around $\ell$ is restored, as restoring just one layer barely shows visible patterns.}
    \label{fig:trace_pythia}
\end{figure}

In the scope of this paper, we only utilize the $26$ \textit{factual} relations from this dataset. We evaluate \textsc{Lre} in Mamba and Pythia for all the $26$ factual relations. We also use this dataset for locating key fact-mediating states in \Cref{sec:locating} and \Cref{app:locate-pythia}. We randomly sample 400 examples $(s, r, o)$ across 6 different factual relations - \textit{place in city}, \textit{country capital city}, \textit{person occupation}, \textit{plays pro sport}, \textit{company hq}, and \textit{product by company}. For each of these examples we randomly select another example within the same relation $(s^*, r, o^*)$ such that $s \neq s^*$ and $o \neq o^*$. The average indirect effect (IE) of applying activation patching over these 400 examples is depicted on Figures \ref{fig:trace_res}, \ref{fig:trace_state}, \ref{fig:sub_last_path} for Mamba-2.8b) and on Figure \ref{fig:trace_pythia} (for Pythia-2.8b). We use the same set of $6$ relations in \Cref{sec:attn_patch} where we adapt attention knock-out to Mamba.

\section{Challenges in Performing Attention Knock-out in Mamba}
\label{app:ablate-retention}

Attention heads in autoregressive transformer LMs and Conv + SSM operations in Mamba play a similar role: bringing/retaining information from the past tokens. Attention \emph{``knock-out''} is a type of causal mediation analysis that tries to understand information flow in transformer LMs by cutting off information propagation from $k^{th}$ token to $q^{th}$ token position. In transformers, each of the attention heads in an attention module $attn^{(\ell)}$ calculates an \textit{attention matrix} $\text{L}$, where $\text{L}_{q,k}$ quantifies how much attention is being paid to the $k^{th}$ token by the $q^{th}$ token with this specific attention head (see \cite{vaswani2017attention} for details on the attention operation). We can block the information flow from $k^{th}$ token to $q^{th}$ token via a specific attention head by simply setting $\text{L}_{q, k} := -\infty$ in the forward pass. 

For Mamba, \cite{ali2024hidden} show that the amount of information retained in the $q^{th}$ token state $s_q^{(\ell)}$, from the \textit{convolved} state at $k^{th}$ token $c_k^{(\ell)}$ (where $k < q$), after the selective-SSM operation (see Equations \ref{eqn:ssm-conv} and \ref{eqn:ssm-end}) can be visualized as an \textit{attention matrix} per channel. Since the selective-SSM operation is linear, the information retained in  $s_q^{(\ell)}$ from $c_k^{(\ell)}$ can be calculated accurately as $\tilde{\alpha}_{q, k}^{(\ell)} = \overline{\text{C}}_q^{(\ell)}\Big(\prod_{i = k+1}^q\overline{\text{A}}_i^{(\ell)}\Big)\overline{\text{B}}_k^{(\ell)} c_k^{(\ell)}$, where $\overline{\text{A}}_i^{(\ell)}$, $\overline{\text{B}}_i^{(\ell)}$, and $\overline{\text{C}}_i^{(\ell)}$ are input-dependent parameters for the $i^{th}$ token. See \cite{gu2023mamba} and \cite{ali2024hidden} for details on selective-SSM operation. 
We ask: can we block the information flow from the $k^{th}$ token to the $q^{th}$ token in Mamba by subtracting out $\tilde{\alpha}_{q, k}^{(\ell)}$ from $s_q^{(\ell)}$? If so, attention knockout experiments in Mamba become feasible. 

We find that blocking information flow via Conv + SSM operation through this specific edge from the $k^{th}$ token to the $q^{th}$ token can be challenging in Mamba. Note that, since $c_k^{(\ell)}$ is a \textit{convolved} state with a receptive field of size 4 in Mamba-2.8b, the states $c_{k+1}^{(\ell)}$, $c_{k+2}^{(\ell)}$, and $c_{k+3}^{(\ell)}$ also retain information from $a_{k}^{(\ell)}$. Which means that even if we subtract $\tilde{\alpha}_{q, k}^{(\ell)}$ from $s_q^{(\ell)}$, these states can \textit{``leak''} information about $a_{k}^{(\ell)}$ to $s_q^{(\ell)}$. To stop this leakage, we would want to subtract from $s_q^{(\ell)}$ all the information retained from $a_k^{(\ell)}$ via $c_{k+1}^{(\ell)}$, $c_{k+2}^{(\ell)}$, and $c_{k+3}^{(\ell)}$ states as well. However, accurately calculating this is challenging because of the $\text{SiLU}$ non-linearity after $\text{Conv1D}$ (see \Cref{eqn:ssm-conv}). 

In our initial experiments we tested subtracting only $\tilde{\alpha}_{q, k}^{(\ell)}$ from $s_q^{(\ell)}$. But we found that Mamba-2.8b could often refer to the $k^{th}$ token from the $q^{th}$ token in copy and factual recall tasks.

\section{Isolating The Contribution of $\text{W}_o^{(\ell)}$}
\label{app:subtract}

\begin{figure}[h!]
    \centering
    \includegraphics[width=\columnwidth]{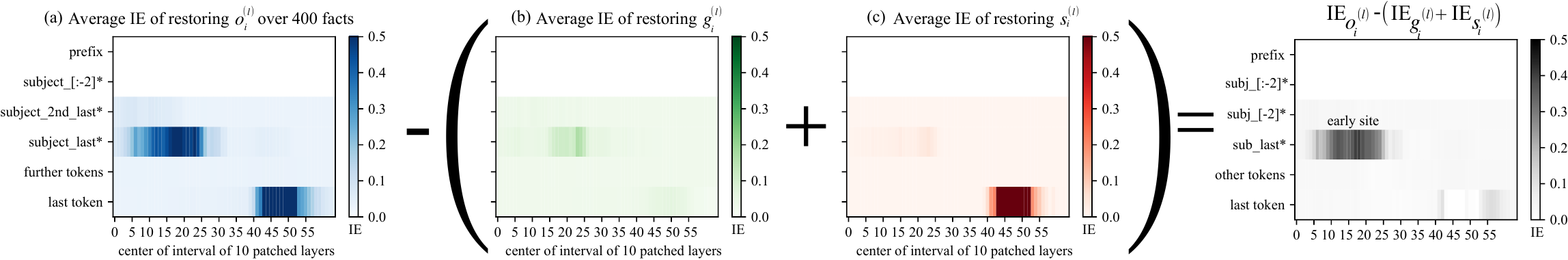}
    \caption{Isolating the contribution of $\text{W}_o^{(\ell)}$. $\text{IE}_{s_i^{(\ell)}} + \text{IE}_{g_i^{(\ell)}}$ subtracted from $\text{IE}_{o_i^{(\ell)}}$. Notice that $\text{IE}_{o_i^{(\ell)}} - \big(\text{IE}_{s_i^{(\ell)}} + \text{IE}_{g_i^{(\ell)}}\big)$ still shows higher causal effect at the early site \big(more pronounced than $\text{IE}_{g_i^{(\ell)}}$\big) while the high causal effect at the late site cancels out.}
    \label{fig:subtract}
\end{figure}

Recall from \Cref{fig:mambablock} and \Cref{eq:block_output} that when $o_i^{(\ell)}$ is restored, the $s_i^{(\ell)}$ and $g_i^{(\ell)}$ are restored as well. To isolate the contribution of only $\text{W}_o^{(\ell)}$ we subtract out $\text{IE}_{s_i^{(\ell)}} + \text{IE}_{g_i^{(\ell)}}$ from $\text{IE}_{o_i^{(\ell)}}$ and plot the results on \Cref{fig:subtract}. Notice that subtracting $\text{IE}_{s_i^{(\ell)}}$ cancels out the high indirect effect at the late site shown by later layers at the last token position. But, together $\text{IE}_{s_i^{(\ell)}} + \text{IE}_{g_i^{(\ell)}}$ cannot cancel out high $\text{IE}_{o_i^{(\ell)}}$ observed at the early site, that is  early-mid layers at the last subject token. This reconfirms the mediating role of $\text{W}_o^{(\ell)}$ at the early site while recalling a fact.  

\section{Locating Key Modules in Pythia-2.8b}
\label{app:locate-pythia}

\begin{figure}[h!]
    \centering
    \includegraphics[width=.95\columnwidth]{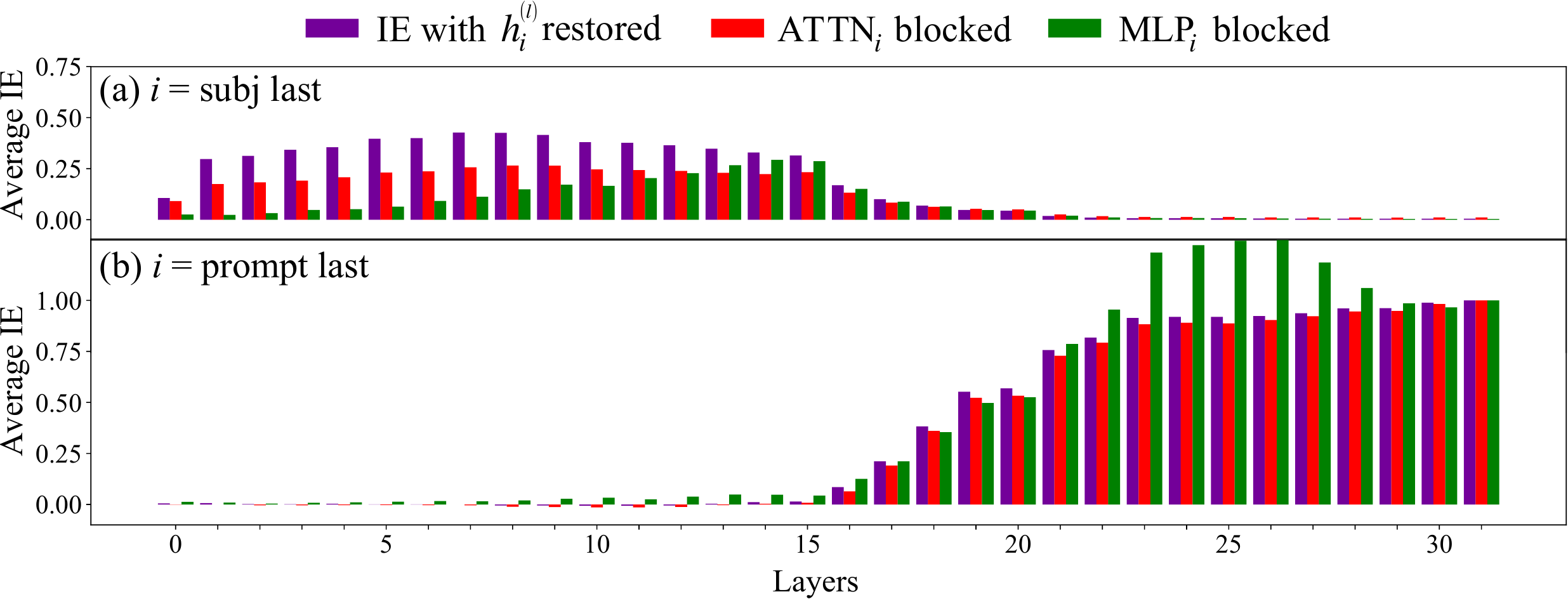}
    \caption{\small Impact of ablating $\text{ATTN}_i$ or $\text{MLP}_i$ on $\text{IE}_{h_i^{(\ell)}}$ for \textbf{(a)} \textit{subject last}  and \textbf{(b)} \textit{prompt last} token positions on Pythia-2.8b}
    \label{fig:path_blocking_pythia}
\vspace{-5pt}
\end{figure}

\newpage
\section{\textsc{Lre} in Pythia-2.8b}
\label{app:lre-pythia}

\begin{figure}[h!]
    \centering
    \includegraphics[width=.95\columnwidth]{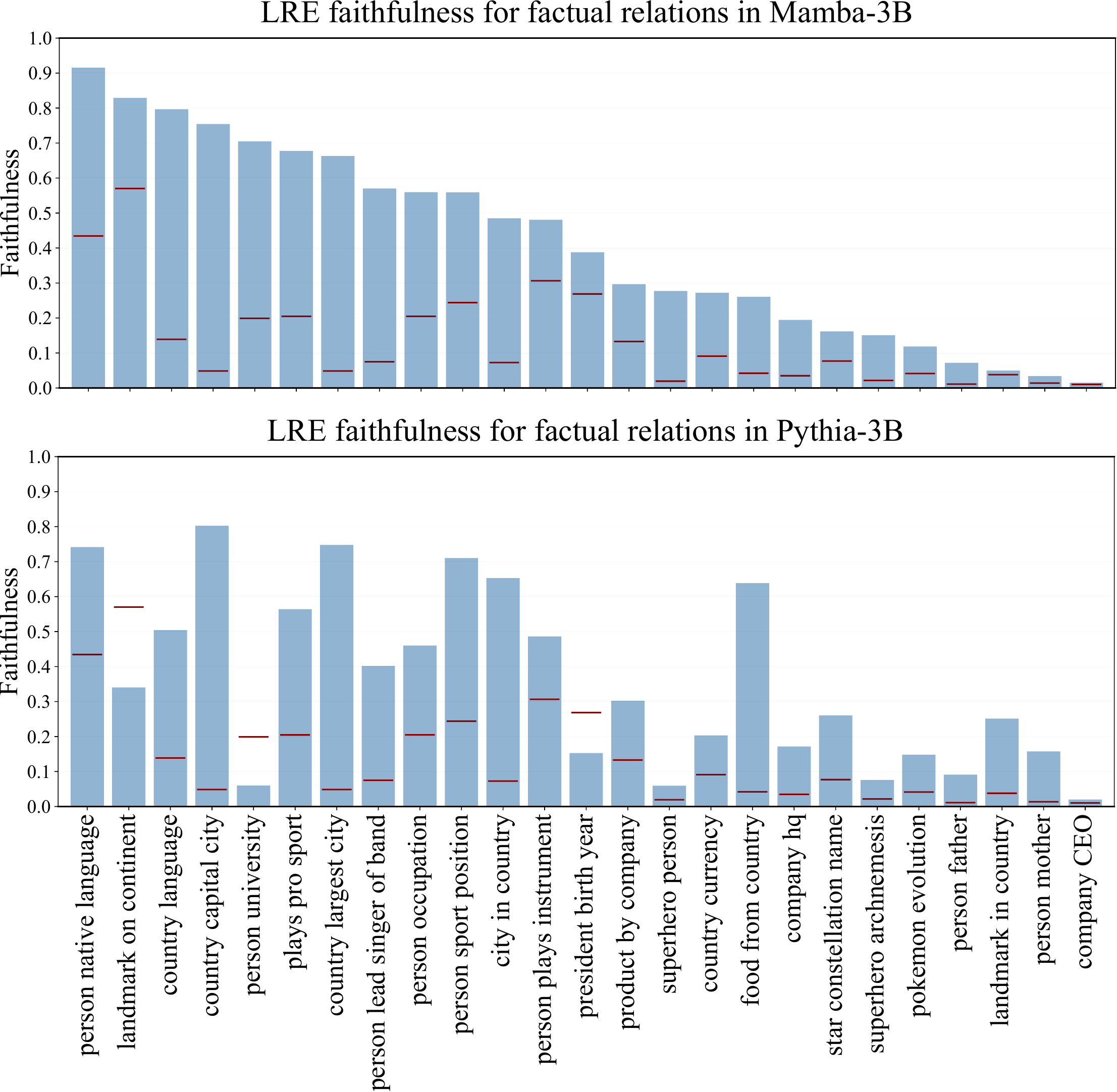}
    \caption{Relation-wise \textsc{Lre} \textit{faithfulness} to the LM decoding function $F$. Horizontal red lines per relation indicate random-choice baseline. We only present results for the \textit{factual} relations in the \textsc{Relations} dataset.}
    \label{fig:lre-faith-compare}
\end{figure}

\newpage
\section{\textsc{Lre} Performance Across Different Relations}
\label{app:lre-relations}

Besides \textit{faithfulness} \cite{hernandez2023linearity} introduced another metric \textit{causality} to measure the performance of \textsc{Lre}. Since \textsc{Lre} is a linear function, it is invertible. Assume that for a fact $(s, r, o)$ \textsc{Lre} can faithfully replace LM computation $F(\mathbf{s}, r)$. Then given the representation $\mathbf{o}^*$ of another object $o$, $\text{J}^{-1} (\mathbf{o}^* - \mathbf{o})$ should give us a $\Delta \mathbf{s}$, such that when added to $\mathbf{s}$, $\tilde{\mathbf{s}} := \mathbf{s} + \Delta\mathbf{s}$, the model computation $F(\tilde{\mathbf{s}}, r)$ should generate $o^*$. See \cite{hernandez2023linearity} for details on this.    

\begin{figure}[h!]
    \centering
    \includegraphics[width=\columnwidth]{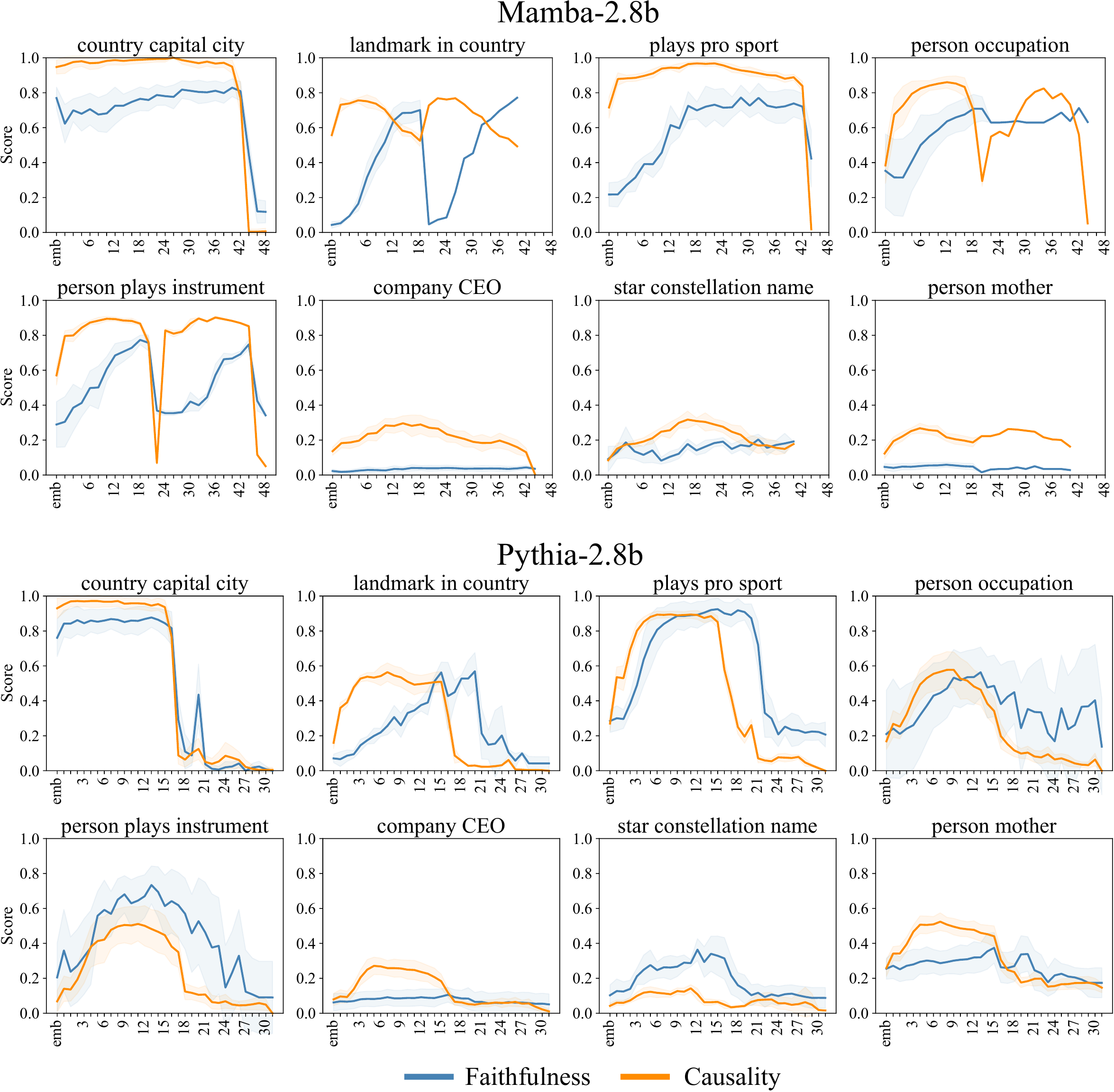}
    \caption{For Mamba, we only perform sweep till layer 48, as \Cref{fig:sub_last_path} suggests negligible activity for later layers at the subject last token}
    \label{fig:lre-sweep-compare}
\end{figure}

\newpage
\section{Activation Patching results on Mamba-2.8b}
\label{app:causal-tracing}

\begin{figure}[h!]
    \centering
    \hspace*{-2.5cm}
    \includegraphics[width=1.35\columnwidth]{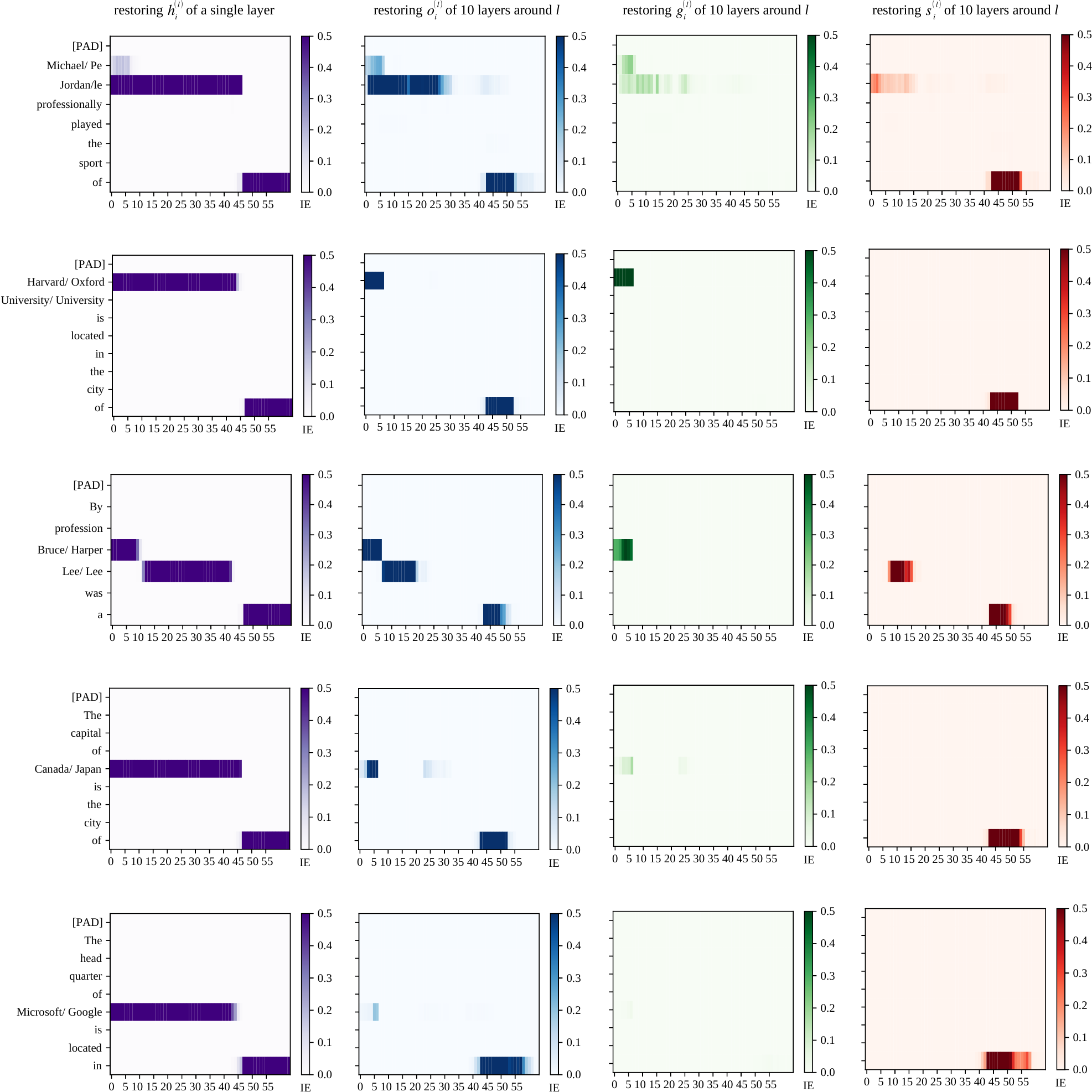}
    \label{fig:causal-tracing-demo}
\end{figure}

\end{document}